\documentclass[10pt, a4paper]{article}
\usepackage{lrec2022} 
\usepackage{multibib}
\newcites{languageresource}{Language Resources}
\usepackage{graphicx}
\usepackage{tabularx}
\usepackage{soul}
\usepackage{bm}
\usepackage{enumitem}
\usepackage{titlesec}
\titleformat{\section}{\normalfont\large\bfseries\center}{\thesection.}{1em}{}
\titleformat{\subsection}{\normalfont\SmallTitleFont\bfseries\raggedright}{\thesubsection.}{1em}{}
\titleformat{\subsubsection}{\normalfont\normalsize\bfseries\raggedright}{\thesubsubsection.}{1em}{}
\renewcommand\thesection{\arabic{section}}
\renewcommand\thesubsection{\thesection.\arabic{subsection}}
\renewcommand\thesubsubsection{\thesubsection.\arabic{subsubsection}}

\usepackage{epstopdf}
\usepackage[utf8]{inputenc}

\usepackage{hyperref}
\usepackage{xstring}

\usepackage{color}

\title{Dynamic Human Evaluation for Relative Model Comparisons}

\name{Thórhildur Thorleiksdóttir\bm{$^{1}$}, Cedric Renggli\bm{$^{1}$}, Nora Hollenstein\bm{$^{2}$}, Ce Zhang\bm{$^{1}$}}

\address{
$^{1}$ETH Zürich, tthorleiks@gmail.com, \{cedric.renggli, ce.zhang\}@inf.ethz.ch\\ 
$^{2}$University of Copenhagen, nora.hollenstein@hum.ku.dk }

\abstract{
Collecting human judgements is currently the most reliable evaluation method for natural language generation systems. Automatic metrics have reported flaws when applied to measure quality aspects of generated text and have been shown to correlate poorly with human judgements. However, human evaluation is time and cost-intensive, and we lack consensus on designing and conducting human evaluation experiments. Thus there is a need for streamlined approaches for efficient collection of human judgements when evaluating natural language generation systems. Therefore, we present a dynamic approach to measure the required number of human annotations when evaluating generated outputs in relative comparison settings. We propose an agent-based framework of human evaluation to assess multiple labelling strategies and methods to decide the better model in a simulation and a crowdsourcing case study. The main results indicate that a decision about the superior model can be made with high probability across different labelling strategies, where assigning a single random worker per task requires the least overall labelling effort and thus the least cost.
\\ \newline  }

\begin{document}

\maketitleabstract

\section{Introduction}
\label{sec:introduction}
Human evaluation is regarded as the primary evaluation metric for natural language generation (NLG) systems due to the lack of automatic metrics that successfully encode quality aspects of generated text~\cite{chaganty-etal-2018-price,celikyilmaz_evaluation_2020}. Still, evaluating systems with human judgements comes with several challenges. Human evaluations are expensive and time-consuming and often demand high-quality judgements (e.g., hiring experts, training non-experts)~\cite{celikyilmaz_evaluation_2020}. A fixed number of annotations can lead to evaluation experiments that are likely to be statistically underpowered to detect the true effects of the corresponding model~\cite{card-etal-2020-little}. Little consensus exists on how to design, conduct, and report human evaluations~\cite{howcroft-etal-2020-twenty}, which has a significant impact on the reliability of the collected human judgements~\cite{novikova-etal-2018-rankme,santhanam-shaikh-2019-towards} and makes it difficult to compare progress across different systems.

When developing NLG systems, automatic metrics are generally applied to track progress~\cite{celikyilmaz_evaluation_2020} despite the fact that common metrics, such as BLEU~\cite{papineni-etal-2002-bleu} and ROUGE~\cite{lin-2004-rouge} have reported flaws and correlate poorly with human judgements~\cite{gehrmann-etal-2021-gem,novikova-etal-2017-need}. These limitations highlight the importance of human evaluation during the development of NLG models to capture the progress of necessary quality aspects of the generated text and to support further improvements for automatic metrics. The ultimate goal is to support the integration of a safe deployment pipeline for NLG models. Once a model with new improvements is estimated to be production-ready, it is evaluated against the current production version with human evaluation and other metrics. Performing human evaluations regularly of different model versions solving the same task can further support the standardisation and confidence in the designed experimental setup and applied evaluation strategies. That process enables streamlining the human evaluation procedure for comparing NLG systems. 

For evaluating generated outputs, we propose a method to dynamically control the required human labelling effort, which denotes the number of needed labels or annotations during an evaluation. The approach supports collecting a sufficient amount of human judgements to make a high probabilistic conclusion when comparing two systems solving the same task. Evaluating models with relative comparison can result in a higher inter-annotator agreement compared to evaluating models independently~\cite{callison-burch_meta-_2007,novikova-etal-2018-rankme}. Thus, with the simplicity of two-alternative forced choice (two-choice) evaluation tasks in mind, we focus on analysing whether we can accumulate sufficient evidence when evaluating two models simultaneously to conclude which model is better with a high probability according to pre-defined text quality criteria.

\paragraph{Contributions}
We define a simple agent-based human evaluation setup to analyse the proposed method of making a high probabilistic decision when comparing two NLG systems with a two-choice evaluation. With our simulation framework, we examine common labelling strategies and the required number of annotations for each strategy with respect to the assigned human capabilities and varying difficulties for model comparisons. Based on findings from the proposed simulation, we then design a case study to evaluate quality aspects of generated outputs from English NLG systems with non-expert human judges in a crowdsourcing setting. With a sufficient amount of collected judgements, we estimate the performance of each labelling strategy and examine the required annotation effort to achieve a confident conclusion of the better model. Our results show that we can make a high probability decision ($0.999$) for all assessed labelling approaches. When comparing the required number of labels for each strategy, assigning a single random worker per request requires the least labelling effort. Moreover, setting different workers per request easily enables parallelization of the evaluation process, which, compared to a sequential evaluation with the same worker, requires less time. We make our code and crowdsourced dataset publicly available.\footnote{\small \url{https://github.com/thorhildurt/dynamic-human-evaluation}}

\section{Related Work} 
\label{sec:related-work}
The importance of analysing and standardising human evaluation methods in text generation tasks has been gaining more attention due to a lack of consensus on how to qualitatively evaluate NLG systems~\cite{van-der-lee-etal-2019-best}. Different task designs and data collection methods (e.g., Likert scales, continuous scales, ranking scales) impact the consistency of collected judgements~\cite{novikova-etal-2018-rankme,santhanam-shaikh-2019-towards}. Comparing different comparison-based data collection methods is beyond the scope of this paper. Our work focuses on analysing the needed labelling effort when comparing two models to make a confident model decision without requiring absolute scores for each model for ranking purposes. There is also an apparent confusion in terminology for evaluating various quality aspects of texts, such as fluency, naturalness, readability, or coherence, to name a few metrics~\cite{howcroft-etal-2020-twenty}. Thus, to reduce confusion when discussing common human evaluation metrics, we generally refer to \emph{quality aspects} of text, but for specific metrics, a clear definition is provided. 

Human evaluations are costly, and thus~\newcite{chaganty-etal-2018-price} focus on reducing the cost by using existing automatic metrics in combination with human evaluation, but only achieve $7$--$13$\% cost reduction compared to performing human evaluation alone. We examine the attained cost in terms of needed labelling effort with respect to our proposed decision method for different human labelling strategies to analyse the difference in cost for equal probabilistic decisions across defined annotation processes (see Section \ref{simulation-results}).

It is becoming increasingly difficult for evaluators to distinguish between machine-generated and human-generated text~\cite{zellers_defending_2020,clark-etal-2021-thats}. Thus, we only compare machine-generated texts and their corresponding quality aspects with respect to each other, without any human-generated reference text. Comparative approaches have proved successful in contrast to direct evaluation~\cite{callison-burch_meta-_2007,novikova-etal-2018-rankme}, but tend to require multiple head-to-head comparisons to achieve statistical significance~\cite{celikyilmaz_evaluation_2020}.~\newcite{sedoc-ungar-2020-item} applied Item Response Theory (IRT) for chatbot evaluation when collecting binary comparisons of system responses to reduce the number of total samples included in the model assessment. ITR can support identifying high-quality sample pairs for measuring the performance from all evaluated response comparisons. Still, the initial collection of human labels is not reduced, and thus neither is the overall labelling effort. 

To standardise NLG evaluation,~\newcite{khashabi_genie_2021} propose a human evaluation leaderboard to automatically track the progress of NLG systems. In contrast, our work does not support an absolute comparison in a leaderboard environment but rather a pairwise relative comparison for choosing the better model. 

\section{Agent-Based Human Evaluation}\label{sec:agent-based-human-eval}
We propose an agent-based simulation of human evaluation for two generative models to analyse the required labelling effort. The following sections describe the proposed simulation framework and configuration for two-choice human evaluation, different labelling strategies, and the proposed decision method for deciding the better model.

\subsection{Simulation Description}\label{subsec:simulation-description}
Simulating two-choice human evaluations gives insights into analysing the required labelling effort when deciding which model is better with a high probability according to pre-defined evaluation categories. A human evaluation generally consists of multiple requests assigned to different human workers. Workers participating in an evaluation typically have varying capabilities, and requests differ in terms of their difficulty, i.e., how hard it is to recognise the correct items according to the task~\cite{whitehill-2009-vote,Vuurens2011HowMS,card-etal-2020-little}.

The configuration of our simulation is inspired by~\newcite{sun-etal-2020-improving}, which provides an analysis of the performance of a novel human annotation method in comparison to standard annotation methods in terms of improving the accuracy for sentence classification. In contrast, our approach does not aim to improve the evaluation accuracy. Instead, it focuses on accumulating sufficient information using different labelling strategies to reach a confident decision with minimum labelling effort when comparing two models simultaneously without pre-existing ground-truth information.

The simulation consists of multiple iterations where workers with varying capabilities evaluate identical requests in each iteration. For each request, a label is assigned depending on the requests' difficulty, the worker's capability, and the labelling strategy in use. Labels are recorded over all requests as well as the required labelling effort according to a given labelling method. For every collected label per request, the proportion of selected labels for each model is updated over an increasing number of requests. 

We assume two generative models, $A$ and $A^{\prime}$ designed to solve the same task. An evaluation consists of $n$ requests pairs $(a_{i}, a_{i}^{\prime})$ sampled from the latent spaces of the two generative models such that $a_{i}\sim z_{A}$ and $a_{i}^{\prime}\sim z_{A^{\prime}}$ where $1 \le i \le n$. We assume that evaluators have sufficient capability when evaluating given request pairs since we do not include evaluators with adversarial behaviour in the simulation. Only one item in each request pair can be selected as the favored item. The evaluation of a single request is an independent action and is not affected by prior events.

\paragraph{Parameters} Every pair $(a_{i}, a_{i}^{\prime})$ has an associated difficulty $d_{i}$ sampled from a difficulty distribution that indicates how hard it is to separate model $A$ with higher target criteria in comparison to model $A^{\prime}$. For the request difficulties we initialise a normal distribution $d\sim \mathcal{N}(\mu,{0.1})$ bounded between $[-1, 1]$. The mean $\mu$ varies between simulations, and the closer the mean is towards $-1$ or $1$, the easier it is to separate the models. When $\mu$ is closer to $-1$ indicates $A^{\prime}$ being better and $A$ is better when $\mu$ is closer to $1$.
The capability $c$ of annotators is sampled from a uniform distribution such that $c\sim Unif(a, b)$ where $a \ge 0$ and $b\le1$. 

For a coherent overview of corner cases and interpretations for evaluator capabilities and request difficulties, we summarise the lowest and highest values when sampling human capabilities and request difficulties for each request pair below:
\begin{itemize}[noitemsep]
    \item \bm{$c=0$}: Incapable annotator. Not fluent in English and does not understand the task. 
    \item \bm{$c=1$}: Highly capable annotator. Fluent in English, strong grammatical skills, understands the task.
    \item \bm{$d=-1$}: Easy to distinguish $a^{\prime}$ as the better item compared to $a$. 
    \item \bm{$d=0$}: Cannot distinguish $a$ being better than $a^{\prime}$ (and vice versa).
    \item \bm{$d=1$}: Easy to distinguish $a$ as the better item compared to $a^{\prime}$.
\end{itemize}

\paragraph{Formulation of the evaluation task}
For simulating the evaluation of any request pair of items sampled from the latent spaces $z_{A}$ and $z_{A^{\prime}}$ with any evaluator, we first compute the product $p=c \cdot d$. The product represents how difficult it is for a worker with capability $c$ to distinguish the better item of any request pair with difficulty $d$. Table~\ref{tab:appendix:corner-cases} provides an overview of the meaning representation of the corner cases for $c$ and $d$ when computing the likelihood of selecting item $a$ or $a^{\prime}$. When a human annotator cannot to distinguish the better item, $c \cdot d = 0$ represents a random selection while $c \cdot d = 1$ or $c \cdot d = -1$ indicate that the correct items according to the given difficulty distribution will be selected.
\begin{table}
\centering
\small
\begin{tabular}{l|l|l|}
\cline{2-3}
                          & $\bm{c=0}$    & $\bm{c=1}$     \\ \hline 
\multicolumn{1}{|l|}{$\bm{d=-1}$} & $c \cdot d=0$ & $c \cdot d=-1$  \\ \hline
\multicolumn{1}{|l|}{$\bm{d=0}$}  & $c \cdot d=0$ & $c \cdot d=0$ \\ \hline
\multicolumn{1}{|l|}{$\bm{d=1}$}  & $c \cdot d=0$ & $c \cdot d=1$  \\ \hline
\end{tabular}
\caption{Computations associated with the meaning representation of the corner cases for human capabilities and request difficulties.}\label{tab:appendix:corner-cases}
\end{table}
We then transform the product from $[-1, 1]$ to a probabilistic range $[0, 1]$ to define the probability $P(a)=(p+1)/2$ of selecting the item generated by model $A$ as the better item, and $P(a^{\prime})=1-P(a)$ for choosing the item generated by $A^{\prime}$. Finally, we abstract the selection with a single Bernoulli trial with $P(1) = P(a)$ and $P(0) = P(a^{\prime})$.

\paragraph{Labelling strategies}
The number of tasks and how many workers are recruited for an evaluation differs across various NLG systems, but best practices suggest to hire at least two or more annotators per task~\cite{van-der-lee-etal-2019-best}. Therefore, we simulate the following strategies in a two-choice evaluation setting:
\begin{itemize}[noitemsep]
    \item \textbf{Fixed Worker}: The same worker is randomly selected to label all requests.
    \item \textbf{One Worker}: A different worker is randomly selected to label each request.
    \item \textbf{N Workers (Majority Vote)}: $N$ workers (randomly selected crowdworkers per request) where each worker labels given request. The majority will decide the final answer for a request where $N$ is an odd number.
    \item \textbf{Max Three Workers}: Each request is randomly assigned to two workers. If they agree on the final answer, then that label is recorded. Otherwise, the request is assigned to one additional worker, which will determine the final answer. 
\end{itemize}

\subsection{Estimating Decision Boundaries}\label{subsec:hoeffding}
The main question we want to answer is when can we decide with high probability which model is better with accumulated evidence according to a given labelling strategy?

Let $X_{i}$ for $1 \leq i \leq n$ be a binary random variable such that $X_{1}, \dots, X_{n}$ represent final answer labels according to given labelling strategy for $n$ requests. When $X_{i}=1$, $a_{i}$ is selected as the better item and when $X_{i}=0$, $a^{\prime}_{i}$ is chosen as the better item. Thus $\overline{X} = \frac{1}{n}\sum_{i=1}^{n} X_{i}$ is the proportion of choices for selecting items generated by model $A$. We assume that when the proportion of selections for a model is $\overline{X} > 0.5$ that we can conclude the better model over $n$ requests. But we want to be able to make such a decision with high probability.  

Concentration inequalities are useful to bound the probability of how far a random variable deviates from its mean. By applying a one-sided version of Hoeffding inequality~\cite{hoeffding_probability_1963}, we can bound the probability $\delta$ with respect to the number of evaluated requests $n$ and error tolerance $t$ such that: 
\begin{equation}
\delta \leq e^{-2nt^{2}}
\end{equation}
The bounded probability $\delta$ represents the likelihood of the sample mean not being included within the given error bound:
\begin{equation}
\delta=P(E[\overline{X}]+t \leq \overline{X})=P(E[\overline{X}]-t \geq \overline{X})
\end{equation}
Accordingly, the probability of the sample mean being within the error bound is $1-\delta$.

The result for a single human evaluation is represented with $\overline{X}$, and with multiple iterations we can further estimate $E[\overline{X}]$ in a simulating setting. But in practice, we lack information regarding $E[\overline{X}]$ when only conducting a single human evaluation. Thus we focus on computing the error bound with respect to the observed sample with sample mean $\overline{X}$, with the same probability: 
\begin{equation}
\delta=P(\overline{X}-t \geq E[\overline{X}])=P(\overline{X}+t \leq E[\overline{X}])
\end{equation}

Thus, when 
$\overline{X} > 0.5$ we make a decision when the corresponding lower bound\footnote{Symmetric computations apply when $\overline{X} < 0.5$ with respect to the upper bound.} satisfies $\overline{X}-t > 0.5$ with $1-\delta$ probability, where:
\begin{equation}
t = \sqrt{-\frac{ln(\delta)}{2n}}
\end{equation}
is computed with fixed $\delta$ for increasing sample size $n$. When we cannot reach a conclusion with sufficiently high probability, the models are indistinguishable according to the accumulated information. 

\section{Simulation Experiments}

We examine the introduced simulation approach to decide upon the better model for all labelling strategies presented in Section~\ref{subsec:simulation-description}. We analyse if strategies that rely on generating a consensus label between several workers (\emph{$N$ Workers with Majority Vote}, \emph{Max $3$ Workers}) gain advantage by requiring fewer request items than having a single worker annotate more requests (\emph{Fixed Worker}, \emph{One Worker}). 

\subsection{Experiment Setup}
We configure the distributions for the simulation parameters as shown in Table~\ref{tab:simulation-configuration}.
To estimate the labelling performance, a single simulation experiment for a given difficulty distribution and human capabilities consists of $1000$ iterations. We simulate the evaluations with three varying difficulty distributions ($d^{(1)}$, $d^{(2)}$, $d^{(3)}$), where all distributions infer, without loss of generality, that model $A$ is better than $A^{\prime}$ over all requests samples. We configure the sample size for each difficulty distribution such that a decision is reachable in each iteration.



\begin{table}
\small
\begin{tabular}{|l|l|c|}
\hline
Parameters         & Distribution   & Sample Size \\ \hline
Workers ($c$)        & $Unif(0.8, 1.0)$ & $100$         \\
Request Diff. ($d^{(1)}$) & $\mathcal{N}(0.25,{0.1})$   & $3500$        \\
Request Diff. ($d^{(2)}$) & $\mathcal{N}(0.125,{0.1})$   & $5000$        \\
Request Diff. ($d^{(3)}$) & $\mathcal{N}(0.0625,{0.1})$  & $15000$       \\ \hline
\end{tabular}
\caption{Parameter configuration for simulation experiments.}\label{tab:simulation-configuration}
\end{table}

\subsection{Selecting the Better Model}
\begin{figure}
\includegraphics[width=\columnwidth]{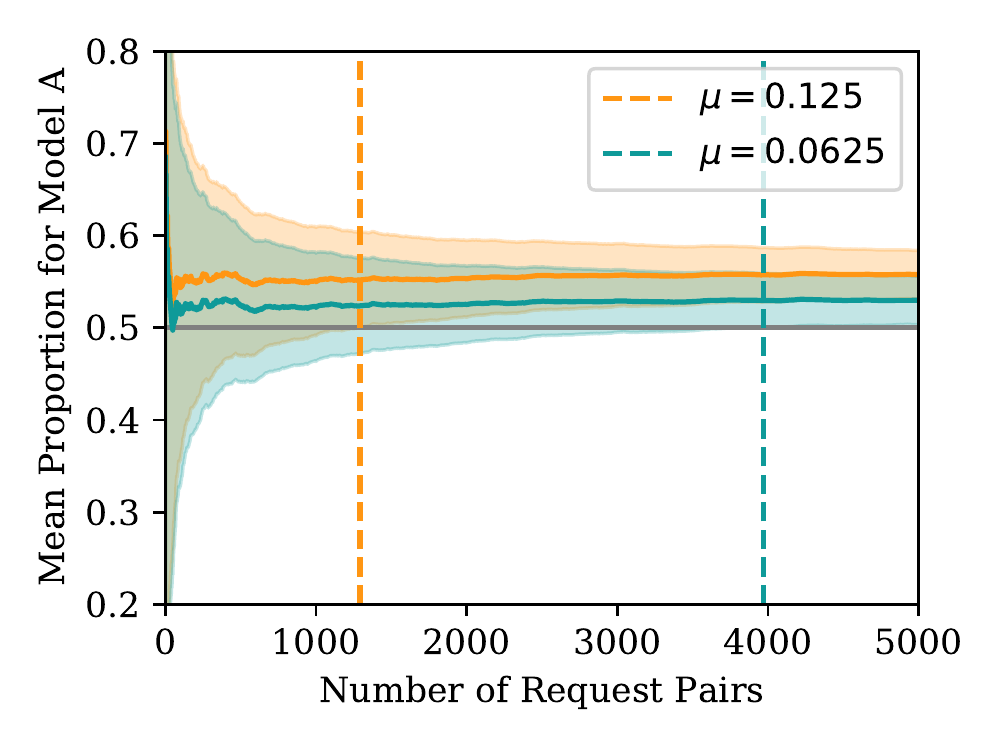}
    \caption{Error bound according to Hoeffding inequality with respect to the proportional mean for selecting model A. The vertical dotted line marks when the lower bound is strictly larger than $0.5$ with $0.999$ probability.}
     \label{fig:Hoeffding}
\end{figure}

Figure~\ref{fig:Hoeffding} shows the corresponding error bounds over increasing number of request pairs for two separate simulations of \emph{One Worker}. For illustration purposes, we visualise the decision process with respect to an estimation of the proportion mean in the simulation setting. The dotted lines indicate the intersection of the lower bound according to the definition of the decision threshold, $\overline{X}-t > 0.5$. The lines show that a decision is achievable with $1-\delta=0.999$ probability in both simulations, but due to differences in request difficulties, the simulation with $\mu=0.0625$ requires roughly $3\times$ more labelling effort than $\mu=0.125$. 

\subsection{Required Labelling Effort}\label{simulation-results}
In each iteration, we compute the proportion of item selections over the number of evaluated request pairs and compute the error bound accordingly. When the computed selection proportion fulfils the defined decision condition with respect to the number of request pairs $n$ and $\delta$, we record the required labelling effort according to the strategy in use. Table~\ref{tab:label-effort-per-method} shows the average required labelling effort over $1000$ iterations to conclude the better model with $0.999$ probability over all labelling strategies for each difficulty distribution. 

As expected, when the difficulty of distinguishing the models increases, the labelling effort increases across all strategies. Since most crowdsourcing platforms assign random workers for each request, it is not common to hire the same worker (the \emph{Fixed Worker} method) to label all tasks. It is possible to design crowdsourcing tasks such that one forces the same worker to label multiple requests to get consistent labelling behaviour over multiple samples~\cite{zhou_hype_2019}. However, forcing the same evaluator to label all requests sequentially is slower than hiring different workers per request (the \emph{One Worker} method), especially when a large sample size is needed for evaluation. Moreover, diverse workers can reduce the impact of having a single worker who is less capable or biased when evaluating all requests. 
The results in Table~\ref{tab:label-effort-per-method} show that we can decide on the better model with $0.999$ probability with fewer workers evaluating more requests, thus requiring less labelling effort compared to assigning $N$ workers per request. According to the bootstrapped confidence intervals, there is no statistical significant difference between the labelling effort required by \emph{Fixed Worker} and \emph{One Worker}. Nevertheless, \emph{One Worker} enables full parallelization, and thus, \emph{One Worker} is a more viable option in comparison to the \emph{Fixed Worker} strategy. Although we assume sufficiently capable workers, we also experimented with increased variance for worker capabilities, which yields similar trends.\footnote{\small \url{https://github.com/thorhildurt/dynamic-human-evaluation/tree/main/paper/supplementary-information}} 

\begin{table}[t]
\small
\centering
\begin{tabular}{|l|c|c|}
\hline
\multicolumn{3}{|c|}{$\mu=0.25$}                   \\ \hline
\textbf{Method}        & \textbf{Avg.}      & \textbf{99\% CI}     \\ \hline
7 Workers     & $866$        & $841-888$     \\
5 Workers     & $722$        & $700-742$     \\
Max 3 Workers & $461$        & $447-476$     \\
Fixed Worker  & \bm{$344$}   & $331-357$     \\
One Worker    & \bm{$338$}   & $325-349$     \\ \hline
\multicolumn{3}{|c|}{$\mu=0.125$}                  \\ \hline
\textbf{Method}        & \textbf{Avg.}      & \textbf{99\% CI}     \\ \hline
7 Workers     & $3647$      & $3536-3767$   \\
5 Workers     & $3141$      & $3040-3236$   \\
Max 3 Workers & $2011$       & $1952-2080$   \\
Fixed Worker  & \bm{$1454$}  & $1404-1502$   \\
One Worker    & \bm{$1440$}  & $1399-1489$   \\ \hline
\multicolumn{3}{|c|}{$\mu=0.0625$}                 \\ \hline
\textbf{Method}      & \textbf{Avg.}      & \textbf{99\% CI}    \\ \hline
7 Workers     & $13302$     & $12965-13609$ \\
5 Workers     & $10850$     & $10607-11114$ \\
Max 3 Workers & $6729$      & $6551-6900$   \\
Fixed Worker  & \bm{$4526$} & $4376-4685$   \\
One Worker    & \bm{$4491$} & $4356-4636$   \\ \hline
\end{tabular}
\caption{Labelling effort for each labelling strategy averaged over $1000$ iterations for three difficulty distributions. A decision is made with $1-\delta=0.999$ probability. Worker capabilities are sampled from $Unif(0.8, 1.0)$. The confidence intervals are computed with bootstrap resampling with $99\%$ confidence.}\label{tab:label-effort-per-method}
\end{table}

\section{Case Study: Evaluating Controlled Text Generation}
Based on the insights of the simulation results, we perform a human
evaluation on NLG models to examine the proposed decision approach for different labelling strategies. First, we introduce the selected NLG models and corresponding configurations to explore different model comparison difficulties. Next, we provide details regarding our human evaluation experimental setup, such as its design, evaluation criteria, and the crowdsourcing setting. Finally, we present our evaluation results.

\subsection{Model Selection}
\label{model-selection}
A common goal for text generation applications is to augment datasets for supervised learning tasks in natural language processing (NLP). The main requirement for these applications is to support controllable text generation that enables systematic control for semantic and syntactic aspects of the generated text.~\newcite{russo-etal-2020-control} recently proposed an NLG model called Control-Generate-Augment (CGA) that learns to control multiple semantic and syntactic attributes of an English sentence with significant performance improvements in downstream NLP tasks.

We perform model comparisons of different difficulty levels for our human evaluation experiments to analyse the changes in required labelling effort between evaluation strategies. For that purpose, we use the CGA framework as a basis for our human evaluation experiments. The publicly available implementation\footnote{\small \url{https://github.com/DS3Lab/control-generate-augment}} enables adjustments to create several variations of attribute-controlled text generation systems of varying quality. To design a simple evaluation of CGA with a concise amount of data, we focus on evaluating sentences of a maximum of 20 tokens rather than evaluating long text paragraphs. 
Table~\ref{tab:cga-model-configuration} provides an overview of the architectural differences between three models trained using the YELP business reviews dataset.\footnote{Dataset retrieved from: \small \url{https://github.com/shentianxiao/language-style-transfer}} The differences between the models are based on modifying the key components required to implement the optimal version of CGA, such as different losses, word-dropout routines and amount of training data.

\begin{table}[t]
\small
\centering
\begin{tabular}{|l|c|c|}
\hline
Model    & WD & Dataset Size         \\ \hline
$L_{ADV}$ + standard WD (V1)       & $0.3$          & $\sim1300$ sent.                \\
$L_{ADV}$ + standard WD (V2)     & $0.7$          & $\sim600.000$ sent.       \\
$L_{CGA}$ + cyclical WD (CGA)  & $\zeta$       & $\sim600.000$ sent.                \\ \hline
\end{tabular}
\caption{The configurations of key components in the CGA framework for three models (WD = word dropout rate).}\label{tab:cga-model-configuration}
\end{table}

The model \emph{$L_{CGA}$ + cyclical WD} is trained using the configuration for the best version of CGA (which we refer to as CGA). The other two models, \emph{$L_{ADV}$ + standard WD (V1)} and \emph{$L_{ADV}$ + standard WD (V2)}, are configured to perform worse in comparison to CGA, i.e., the generated sentences from V1 and V2 are expected to perform worse concerning standard quality aspects such as grammatical correctness, or naturalness. The two models differ vastly in performance mainly due to different amounts of data used during training, since V1 is only trained with a $~0.2\%$ of the available data. Table~\ref{tab:attribute-matching} provides an overview of automatic evaluation with attribute matching for each model as described in \cite{russo-etal-2020-control}, which shows the performance impact of each model configuration. 

\begin{table}
\centering
\small
\begin{tabular}{|l|ccc|}
\hline
Model & Sentiment & Tense   & Person  \\ \hline
V1    & 65.60\%   & 39.48\% & 41.03\% \\
V2    & 95.93\%   & 96.53\% & 56.53\% \\
CGA   & 98.68\%   & 98.08\% & 56.02\% \\ \hline
\end{tabular}
\caption{Attribute matching accuracy (in \%) of $6000$ generated sentences for each evaluated model.}
\label{tab:attribute-matching}
\end{table}

The $L_{ADV}$ loss combines variational autoencoder (VAE) and discriminator loss functions, while the $L_{CGA}$ loss is $L_{ADV}$ combined with a context-aware loss. The standard WD is based on the dropout routine applied in~\cite{bowman-etal-2016-generating}, which uses a fixed word dropout rate. In contrast,~\newcite{russo-etal-2020-control} apply cyclical WD, which represents a cyclical-dropout routine to compute the dropout rate $\zeta$ in each training step.
The detailed architecture components of CGA, e.g., the formal definition of the loss functions and the cyclical dropout routine, can be found in~\cite{russo-etal-2020-control}. 

We construct two settings of model comparisons based on empirical observations and automatic evaluation: 
(1) The \emph{major improvements} setting refers to the comparison between CGA and V1, where the differences are easily distinguishable; and (2) the \emph{minor improvements} setting, a more challenging comparison where CGA is compared to V2. In both cases, the models are expected to be distinguishable.

\subsection{Evaluation Criteria}
The models generate sentences by controlling semantic and syntactic attributes, and thus it is important to evaluate whether the sentences include the provided attributes. Pre-trained classifiers can be used for automatic evaluation of attribute matching, but it remains difficult to automatically capture quality aspects of the generated text~\cite{hashimoto-etal-2019-unifying}. Therefore, we focus on evaluating quality aspects of the generated sentences for our human evaluation experiment since the models must generate natural, grammatically correct and coherent sentences. 
Moreover,~\newcite{van-der-lee-etal-2019-best} emphasise the importance of focusing on a single quality criterion per evaluation and thus, we focus on one specific evaluation criterion in our experiments. We evaluate the \textbf{naturalness} of the generated sentences, representing whether a given text could have been produced by a native speaker~\cite{novikova-etal-2018-rankme}.

\subsection{Data}
As mentioned before, we train three variations of attribute control models for our human evaluation experiments. We generate $6000$ sentences for each model by controlling combinations of the three following attributes: verb tense, sentiment, and person number. When preparing the pairwise comparison, we pair sentences from each model on matching attributes and similar sentence length to reduce annotator bias towards shorter or longer sentences. Duplicate sentences are removed from each sample of generated sentences. 
The order of the sentences in each pair is randomised as well as the list containing all sentence pairs. From each list of sentence pairs, we sample $500$ pairs which are then published as evaluation requests in a crowdsourcing setting for each experiment.

\subsection{Crowdsourcing Setting}
\begin{figure*}
    \centering
    \includegraphics[width=0.85\linewidth]{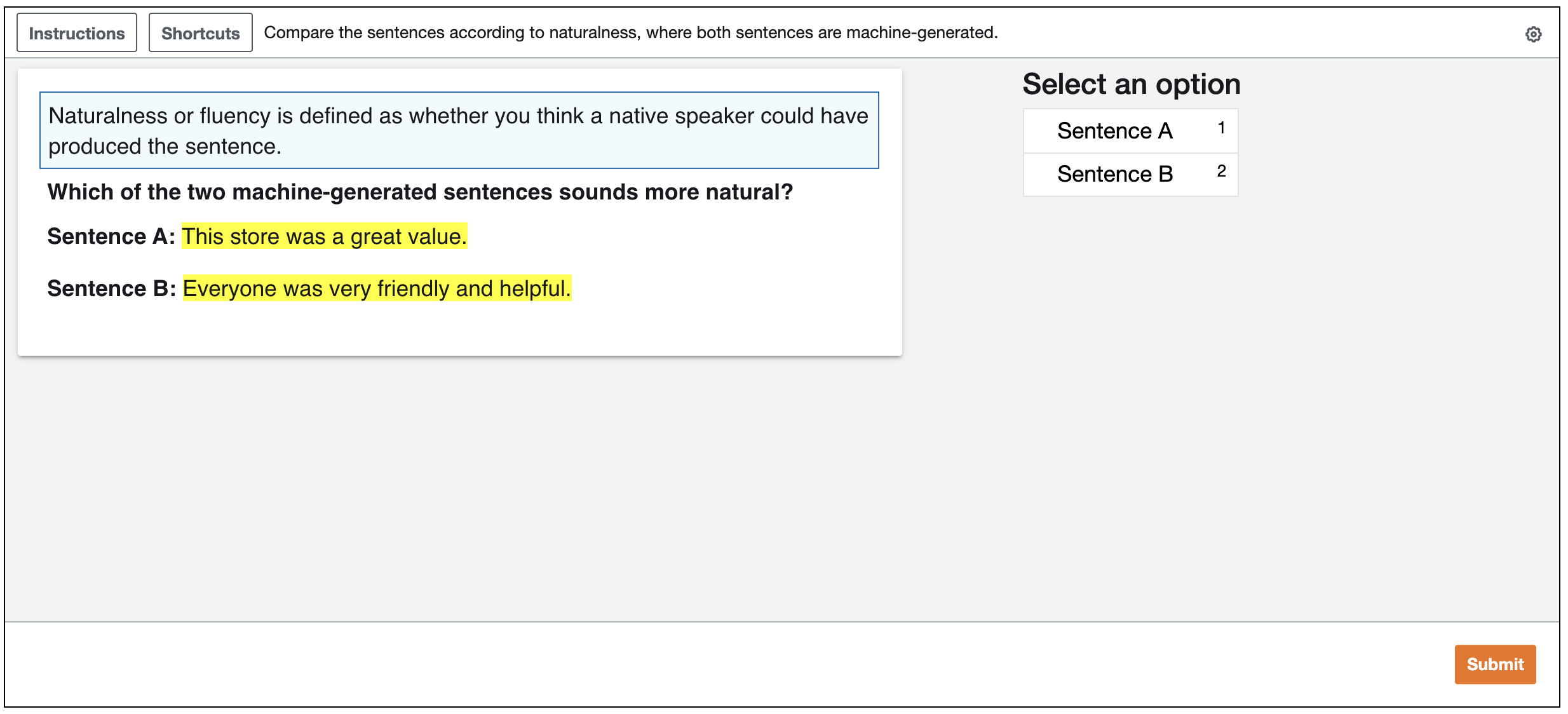}
    \caption{The task interface on Amazon Mechanical Turk.}\label{task-design}
\end{figure*}

We use Amazon Mechanical Turk (AMT) to conduct the human evaluation, a popular platform to collect evaluation from non-expert annotators for various NLG tasks~\cite{celikyilmaz_evaluation_2020}. 
Figure~\ref{task-design} represents the design of the evaluation task on AMT. To maintain clarity throughout the evaluation, we include the definition of the evaluation criteria in each task. We highlight the machine-generated sentences to assist the workers to focus on the central part of the task in contrast to additional information present in the interface.
To maintain quality control, we further increase the recommendation provided by~\newcite{Berinsky2012EvaluatingOL} on qualification requirement on AMT for more reliable worker performance. All workers must have at least $10.000$ approved HITs and an approval rate $\geq98\%$. To ensure familiarity with the English language, the location of the workers is required to be in the US or GB. In each experiment, we collected $5000$ evaluations for $500$ sentence pairs each evaluated by $10$ random workers, where the payment per comparison is $\$0.02$. 

To fairly compensate workers, we estimated the hourly wage to succeed the federal minimum wage in the US. The expected wage was \$9/hour, where the average time per task was estimated to be 8 seconds. All workers that met the qualification requirements and participated in the evaluation were paid independent of their provided answers. Anonymity is preserved for the collected annotations where participants privacy rights are respected. 


\begin{table*}
\small
\centering
\begin{tabular}{|l|cc|cc|cc|}
\hline
\textbf{V1 vs CGA}  & \multicolumn{2}{c|}{$1-\delta=0.99$} & \multicolumn{2}{c|}{$1-\delta=0.999$} & \multicolumn{2}{c|}{$1-\delta=0.9999$} \\ \hline
Method         & Avg.              & 99\% CI         & Avg.                & 99\% CI         & Avg.                & 99\% CI          \\ \hline
7 Workers      & $70$         & 70--70          & $98$          & 98--98          & $133$         & 133--133         \\
5 Workers      & $50$         & 50--50          & $70$          & 70--70          & $95$          & 95--95           \\
Max 3 Workers  & $30$         & 30--30          & $43$          & 42--44          & $59$          & 58--60           \\
One Worker & \bm{$11$}        & 10--12          & \bm{$19$}    & 17--20          & \bm{$26$}     & 25--28           \\ \hline
\multicolumn{1}{l}{}        & \multicolumn{1}{l}{}          & \multicolumn{1}{l}{} & \multicolumn{1}{l}{}          & \multicolumn{1}{l}{}                   & \multicolumn{1}{l}{}            \\ \hline
\textbf{V2 vs CGA (R1)} & \multicolumn{2}{c|}{$1-\delta=0.99$} & \multicolumn{2}{c|}{$1-\delta=0.999$} & \multicolumn{2}{c|}{$1-\delta=0.9999$} \\ \hline
Method         & Avg.                 & 99\% CI       & Avg.                 & 99\% CI        & Avg.                 & 99\% CI         \\ \hline
7 Workers      & $501$        & 436--576      & $823$        & 750--902       & $1235$       & 1131--1340      \\
5 Workers      & $452$        & 405--505      & $671$        & 604--736       & $1010$       & 917--1102       \\
Max 3 Workers  & \bm{$327$}       & 283--367      & $519$        & 466--566       & $701$        & 647--760        \\
One Worker & \bm{$273$}*       & 244--302      & \bm{$356$}*        & 326--380       & \bm{$385$}*        & 350--413        \\ \hline
\multicolumn{1}{l}{}        & \multicolumn{1}{l}{}          & \multicolumn{1}{l}{} & \multicolumn{1}{l}{}          & \multicolumn{1}{l}{}                   & \multicolumn{1}{l}{}            \\ \hline
\textbf{V2 vs CGA (R2)} & \multicolumn{2}{c|}{$1-\delta=0.99$} & \multicolumn{2}{c|}{$1-\delta=0.999$} & \multicolumn{2}{c|}{$1-\delta=0.9999$} \\ \hline
Method         & Avg.                 & 99\% CI       & Avg.                 & 99\% CI        & Avg.                 & 99\% CI         \\ \hline
7 Workers      & $509$          & 432--590    & $954$         & 903--1005     & $1266$         & 1200--1333    \\
5 Workers      & $357$          & 306--408    & $668$         & 622--724      & $987$          & 927--1042     \\
Max 3 Workers  & \bm{$240$}     & 207--271    & $430$         & 391--473      & $603$          & 560--647      \\
One Worker & \bm{$187$}      & 166--211    & \bm{$281$}     & 259--303      & \bm{$350$}*     & 331--371      \\ \hline
\end{tabular}
\caption{Labelling effort for comparing V1 vs CGA (top), V2 vs CGA (R1) (middle) and V2 vs CGA (R2) (bottom). Required labelling effort per decision is averaged over $100$ iterations for increasing probability. The confidence intervals are computed with bootstrap resampling with $99\%$ confidence. * marks that deciding the better model was not achieved in some iterations with the collected sample of evaluated request pairs.}\label{tab:label-effort-per-method-mturk}
\end{table*}

\subsection{Human Evaluations}
We aim to analyse the required labelling effort for different labelling strategies for varying request difficulties with the model comparison settings introduced in Section~\ref{model-selection}. For the more challenging comparison (V2 vs CGA), the experiment was repeated (R1, R2) on two distinct days to analyse the reliability of the labelling effort results in a crowdsourcing setting across time and workers. 

To analyse the collected human evaluations, we conduct a procedure similar to our simulation method to better represent the underlying distribution. We run $100$ iterations over the request pairs evaluated on AMT. We sample random workers without replacement for a single request in each iteration, depending on the given evaluation method. Note, that due to the randomness present in the worker selection on AMT, there is no guarantee that the same worker evaluates multiple tasks. Therefore, we omit the \emph{Fixed Worker} method from our analysis with real human data. Thus, based on the observed two-choice sentence comparisons, we randomly sample 100 human evaluations for identical request pairs.

\subsection{Results}
Overall, we reach the same decision as in the simulation experiments on selecting the better model in a two-choice setting with a high probability ($0.9999$) for all labelling strategies. The \emph{One Worker} labelling strategy requires the least labelling effort when performing model comparisons for both major and minor improvements.

\paragraph{Major Improvements}
An overview of the labelling effort required for a decision with increasing probability $1-\delta$ when comparing CGA and V1 is presented in Table \ref{tab:label-effort-per-method-mturk} (top). The comparison between CGA and V1 yields a high consensus amongst evaluators shown with high agreement score (Fleiss $\kappa = 0.69$) and minimal variation in the required labelling effort, especially for the methods which require several evaluators per request. That further indicates that there exists a common understanding of the goal of the designed evaluation task. 

The majority voting methods with $N=5$ and $N=7$ workers result in consistent labelling effort over the increasing probabilities of $0.99$, $0.999$, and $0.9999$. However, despite the consistency in the required labels, assigning each request to random worker (\emph{One Worker}) yields the least required labelling effort to conclude the better model over $100$ sampled human evaluations. 

\begin{table}
\small
\centering
\begin{tabular}{|l|ccc|}
\hline
 Repetition  & $0.99$ & $0.999$ & $0.9999$ \\ \hline
R1 & 95\%            & 80\%             & 49\%              \\
R2 & 100\%           & 100\%            & 96\%              \\ \hline
\end{tabular}
\caption{The ratio (\%) for how often a decision is made for a sample of the evaluated requests pairs over $100$ iterations with varying probability for two repetitions of V2 vs CGA.}\label{tab:decision-count-for-random-worker}
\end{table}

\paragraph{Minor Improvements}
With a more challenging comparison, where the request pairs contain potentially more ambiguous human selections, it is expected that the average required labelling effort increases compared to easier comparison tasks as examined in Section~\ref{simulation-results}. Lower annotator agreement scores show less consensus for both repetitions than in the previous setting, where R1 results in Fleiss $\kappa=0.27$ and R2 in Fleiss $\kappa=0.38$. The required labelling effort over increasing probability for V2 vs CGA (R1) is summarised in Table~\ref{tab:label-effort-per-method-mturk} (middle) and for V2 vs CGA (R2) in Table~\ref{tab:label-effort-per-method-mturk} (bottom). 
The experiment setup of R1 and R2 are identical, the only difference being the time and date of the evaluation.

Similar to previous findings, \emph{One Worker} requires less labelling effort in comparison to all labelling methods in both R1 and R2. For a decision with $0.99$ probability, there is no statistically significant difference between the required labelling effort between \emph{Max 3 Workers} and \emph{One Worker} in both R1 and R2 for the computed confidence intervals. However, with increasing probability \emph{One Worker} requires significantly less labelling effort in comparison to all methods in both R1 and R2. 

Table~\ref{tab:decision-count-for-random-worker} shows the ratio of how often we achieve a decision with the \emph{One Worker} strategy over sampled evaluated request pairs. The ratio is 1 for all other evaluation methods on the given data. In R1, this ratio decreases with increasing probability since the method requires more request pairs beyond the provided data when aiming for more confident decisions. 


\subsection{Discussion}
Performing a simulated human evaluation has no cost, which is a clear advantage over performing an actual human evaluation. But the simulation only relies on standard probability distributions. It thus does not reflect complex human features that can impact a decision, which emphasizes the importance of additionally conducting a real human evaluation. But simulating different labelling strategies can further support the human evaluation design and development of novel labelling strategies that require less cost. The results produced with simulated and crowdsourced human evaluations show similar trends. In both experiments, assigning a single worker to requests results in less labelling effort for a confident decision compared to multiple workers per request. Even though assigning a single worker per request pair goes against the evaluation recommendations provided by~\newcite{van-der-lee-etal-2019-best}, 
\newcite{khashabi_genie_2021} found that collecting one label per instance resulted in less variance when computing leaderboard scores with human evaluations in contrast to collecting multiple labels per instance. Our findings are in line with~\newcite{khashabi_genie_2021} and show that requiring a single label per request over a sufficient number of requests yields the same decision with the same probability as requiring multiple labels per request but needing less labelling effort and thus less cost.

\section{Conclusion and Future Work}
We propose a method of human evaluation to conclude with high probability which of two models is better in an agent-based simulation and crowdsourcing setting. Our approach achieves confident decisions for all analysed labelling strategies with varying worker capabilities and request difficulties. When comparing the required effort across different labelling methods, using a single worker per request is the strategy that requires the least labelling effort. Moreover, assigning a different worker per request enables trivial parallelisation such that less time is required for evaluation.

The current decision method is a viable first option to analyse the needed labelling efforts across different labelling strategies. As part of potential future work, other approaches beyond Hoeffding inequality can be explored to infer the stopping criteria. Additionally, in the current simulation setting, we cannot estimate how many labels are required for a real human evaluation nor conclude whether two models are indistinguishable according to a given evaluation criteria, which can be further analysed in future work. We hope the proposed method will enable the design of improved evaluation strategies to require fewer samples evaluated by humans to select the better NLG model with high confidence.

\section{Acknowledgements}
CZ and the DS3Lab gratefully acknowledge the support from the Swiss State Secretariat for Education, Research and Innovation (SERI)’s Backup Funding Scheme for European Research Council (ERC) Starting Grant TRIDENT (101042665), the Swiss National Science Foundation (Project Number 200021\_184628, and 197485), Innosuisse/SNF BRIDGE Discovery (Project Number 40B2-0\_187132), European Union Horizon 2020 Research and Innovation Programme (DAPHNE, 957407), Botnar Research Centre for Child Health, Swiss Data Science Center, Alibaba, Cisco, eBay, Google Focused Research Awards, Kuaishou Inc., Oracle Labs, Zurich Insurance, and the Department of Computer Science at ETH Zurich.

\clearpage
\section{Bibliographical References}\label{reference}

\bibliographystyle{lrec2022-bib}
\bibliography{lrec2022, anthology}


\end{document}